\documentclass[final]{l4dc2026}

\title[Learning Quantized Continuous Controllers for Integer Hardware]{Learning Quantized Continuous Controllers for Integer Hardware}

\usepackage{times}
\usepackage{tikz}
\usepackage{pgfplots}
\usepackage{amsmath}
\usepackage{amssymb}
\usepackage{mathtools}
\usepackage{wrapfig}
\usepackage{comment}
\usepackage{color}
\usepackage{xcolor}
\usepackage{hyperref}
\usepackage{subcaption}
\usepackage{graphicx}
\usepackage{siunitx}
\usepackage{threeparttable}
\usepackage{multicol}
\usepackage{float}
\usepackage{multirow}
\usepackage{booktabs}
\usetikzlibrary{intersections}
\usetikzlibrary{calc}
\usepackage{xstring}

\pgfplotsset{compat=1.18}
\usepgfplotslibrary{fillbetween,groupplots}

\usepackage{xspace}
\makeatletter
\DeclareRobustCommand\onedot{\futurelet\@let@token\@onedot}
\def\@onedot{\ifx\@let@token.\else.\null\fi\xspace}

\newcommand{\QDQ}{\ensuremath{\mathrm{QDQ}}\xspace}
\newcommand{\Q}{\ensuremath{\mathrm{Q}}\xspace}

\definecolor{istaDark}{HTML}{005844}

\definecolor{istaDark}{HTML}{005844}
\definecolor{istaMid}{HTML}{2B8A70}

\def\eg{{e.g}\onedot} 
\def\ie{{i.e}\onedot} 
 
 \def\vs{{vs}\onedot}

\newcommand{\ms}{\ensuremath{\mathrm{ms}}}
\newcommand{\mus}{\ensuremath{\mu\mathrm{s}}}

\pgfplotsset{
  myaxis/.style={
    width=7.2cm, height=4.0cm,
    xmin=2, xmax=8,
    xtick={8,7,6,5,4,3,2},
    x dir=reverse, 
    scaled y ticks=false,
    xlabel={}, ylabel={Reward},
    ymajorgrids=true,
    legend cell align={left},
    error bars/y explicit,
    error bars/error bar style={line width=0.7pt},
    error bars/error mark options={line width=0.7pt, rotate=90, mark size=1.8pt},
  },
  sizeaxis/.style={
    width=3.4cm, height=3.4cm,
    x dir=reverse, 
    xlabel={}, ylabel={Reward},
    ymajorgrids=true,
    legend cell align={left},
    error bars/y explicit,
    error bars/error bar style={line width=0.7pt},
    error bars/error mark options={line width=0.7pt, rotate=90, mark size=1.8pt},
  },
  stdaxis/.style={
    width=3.4cm, height=3.4cm, 
    xlabel={}, 
    ylabel={Reward [$\times 1000$]},
    yticklabel={\pgfmathparse{\tick/1000}\pgfmathprintnumber{\pgfmathresult}}, ymajorgrids=true,
    error bars/y explicit,
    error bars/error bar style={line width=0.7pt},
    error bars/error mark options={line width=0.7pt, rotate=90, mark size=1.8pt},
  },
timeaxis/.style={
    width=5.cm, height=4.0cm,
    grid=both,
    xlabel={Enviroment Step},
    title style={font=\small},
    label style={font=\small},
    scaled y ticks=false,
    legend style={font=\footnotesize},
    line width=0.9pt,
    tick style={line width=0.6pt},
    scaled ticks=false,
    xmin=0, xmax=1000000,
    xtick={0,500000,1000000},
    xticklabels={0,0.5M,1M},
  },
  fpstyle/.style={line width=1.4pt, dashed, color=black!55},
  fpband/.style={draw=none, fill opacity=0.15},
     qA/.style={
      line width=1.2pt,
      color={rgb,1:red,0.1216;green,0.4667;blue,0.7059}, %
      mark=*,
      mark size=2.0pt,
      mark options={solid},     %
    },
    qB/.style={
      line width=1.2pt,
      color={rgb,1:red,1.0;green,0.498;blue,0.0549},      %
      mark=triangle*,
      mark size=2.2pt,
      mark options={solid},
    },
    qC/.style={
      line width=1.2pt,
      color={rgb,1:red,0.1725;green,0.6275;blue,0.1725},  %
      mark=square*,
      mark size=2.0pt,
      mark options={solid},
    },
    qD/.style={
      line width=1.2pt,
      color={rgb,1:red,0.6275;green,0.1;blue,0.1},  %
      mark=star,
      mark size=3.0pt,
      mark options={solid},
    },
    qAN/.style={
      line width=1.2pt,
      color={rgb,1:red,0.1216;green,0.4667;blue,0.7059}, %
      mark=None,
      mark size=2.0pt,
      mark options={solid},     %
    },
    qBN/.style={
      line width=1.2pt,
      color={rgb,1:red,1.0;green,0.498;blue,0.0549},      %
      mark=None,
      mark size=2.2pt,
      mark options={solid},
    },
    qCN/.style={
      line width=1.2pt,
      color={rgb,1:red,0.1725;green,0.6275;blue,0.1725},  %
      mark=None,
      mark size=2.0pt,
      mark options={solid},
    },
}
\tikzset{
  fpband/.style={fill opacity=0.15, draw=none, fill=black},
  qAband/.style={fill opacity=0.15, draw=none, fill=blue},
  qBband/.style={fill opacity=0.15, draw=none, fill=red},
  qCband/.style={fill opacity=0.15, draw=none, fill=orange},
}

\definecolor{sz512}{RGB}{217,95,2}   %

\newcommand{\envpanel}[2]{%
  \addplot[name path=fp_up, draw=none, forget plot]
    table[col sep=comma, x=global_step, y expr=\thisrow{mean}+\thisrow{std}]
      {#1/#2/fp.csv};
  \addplot[name path=fp_lo, draw=none, forget plot]
    table[col sep=comma, x=global_step, y expr=\thisrow{mean}-\thisrow{std}]
      {#1/#2/fp.csv};
  \addplot[fpband, forget plot] fill between[of=fp_up and fp_lo];
  \addplot[fpstyle]
    table[col sep=comma, x=global_step, y=mean]
      {#1/#2/fp.csv};
  \addlegendentry{FP32}

  \addplot+[
    qA, error bars/.cd, y dir=both, y explicit,
  ] table[col sep=comma, x=global_step, y=mean, y error=std]
    {#1/#2/input.csv};
  \addlegendentry{input}

  \addplot+[
    qB, error bars/.cd, y dir=both, y explicit,
  ] table[col sep=comma, x=global_step, y=mean, y error=std]
    {#1/#2/weight.csv};
  \addlegendentry{core}

  \addplot+[
    qC, error bars/.cd, y dir=both, y explicit,
  ] table[col sep=comma, x=global_step, y=mean, y error=std]
    {#1/#2/output.csv};
  \addlegendentry{output}
   
   \addplot+[
    qD, error bars/.cd, y dir=both, y explicit,
  ] table[col sep=comma, x=global_step, y=mean, y error=std]
    {#1/#2/all.csv};
  \addlegendentry{all}
}

\newcommand{\plotbandmean}[4]{%
  \addplot[name path=#2_up, draw=none, forget plot]
    table[col sep=comma, x=global_step, y expr=\thisrow{mean}+\thisrow{std}]{#1};
  \addplot[name path=#2_lo, draw=none, forget plot]
    table[col sep=comma, x=global_step, y expr=\thisrow{mean}-\thisrow{std}]{#1};
  \addplot[#4, forget plot] fill between[of=#2_up and #2_lo];
  \addplot[#3]
    table[col sep=comma, x=global_step, y=mean]{#1};
}
\newif\ifshowlegend

\newcommand{\bitpanel}[3]{%
  \plotbandmean{#1/#2/fp.csv}{fp}{fpstyle}{fpband}
  \addlegendentry{FP32}

  \plotbandmean{#1/#2/bits-#3.csv}{b2}{qAN}{qAband}
 \addlegendentry{Quantized Model} %

}

\newcommand{\sizepanel}[3]{%
  \plotbandmean{#1/#2/fp.csv}{fp}{fpstyle}{fpband}
  \ifshowlegend\addlegendentry{FP32}\fi

  \plotbandmean{#1/#2/size_8_#3_8.csv}{b2}{qA}{qAband}
  \ifshowlegend\addlegendentry{Size}\fi

}

\newcommand{\inputpanel}[4]{%
  \plotbandmean{#1/#2/fp.csv}{fp}{fpstyle}{fpband}
  \ifshowlegend\addlegendentry{FP32}\fi

  \plotbandmean{#1/#2/#3_8_#4_8.csv}{b2}{qA}{qAband}
  \ifshowlegend\addlegendentry{Size}\fi

}

\newcommand{\noisepanel}[2]{%
  \plotbandmean{#1/#2/fp32_values.csv}{fp}{fpstyle}{fpband}
  \addlegendentry{FP32}
    \plotbandmean{#1/#2/noise_levels_varied.csv}{b3}{qB}{qBband}
  \addlegendentry{Quantized Model} %

}

\usepackage{booktabs, multirow, graphicx, siunitx}

\newcommand{\ns}[1]{\SI{#1}{\nano\second}}
\newcommand{\numk}[1]{\SI[scientific-notation=fixed,fixed-exponent=3]{#1}{k}}

 \coltauthor{\Name{Fabian Kresse} \Email{fabian.kresse@ist.ac.at}\\
  \Name{Christoph H. Lampert} \Email{chl@ist.ac.at}\\
  \addr Institute of Science and Technology Austria (ISTA), Am Campus 1, 3400 Klosterneuburg, Austria}

\begin{document}

\maketitle

\begin{abstract}%
Deploying continuous-control reinforcement learning policies on embedded hardware requires meeting tight latency and power budgets.  Small FPGAs can deliver these, but only if costly floating-point pipelines are avoided. 
We study quantization-aware training (QAT) of policies for integer inference and we present a learning-to-hardware pipeline that automatically selects low-bit policies and synthesizes them to an Artix-7 FPGA. %
Across five MuJoCo tasks, we obtain policy networks that are competitive with full precision (FP32) policies but require as few as 3 or even only 2 bits per weight, and per internal activation value, as long as input precision is 
chosen carefully. 
On the target hardware, the selected policies achieve inference latencies on the order of microseconds and consume microjoules per action, favorably comparing to a quantized reference.
Last, we observe that the quantized policies exhibit increased 
input noise robustness compared to the floating-point baseline. 
\end{abstract}

\medskip

\section{Introduction}

Deep Reinforcement Learning (RL) has achieved strong results in continuous control with real-time requirements, such as robotic manipulation~\citep{kalashnikov2018scalable}, unmanned aerial vehicle (UAV) control~\citep{kaufmann2023champion}, and fusion plasma control~\citep{degrave2022magnetic}.

Beyond \emph{high control quality}, practical deployment of such systems hinges on \emph{latency}, \emph{energy}, and \emph{compatibility with the available hardware}: Constraints on \emph{latency}, \ie the sensor-to-actuator delay, usually determine which tasks are feasible: 
for autonomous-driving tasks, latency commonly has to stay below \(100\,\ms\), even at modest driving speeds~\citep{kato2015open}. For the inner loop of a robotic arm or quadcopter control, a latency of %
only a few milliseconds can be required~\citep{zhang2020modular,dimitrova2020towards}.
For special-purpose settings, such as nuclear fusion, controllers have to operate even faster, allowing for a latency of at most \(100\,\mus\) per decision \citep{degrave2022magnetic}.
Note that inference is only one piece of the sense-act pipeline. Therefore, the actual compute budget available to the processing unit is even smaller. Furthermore, a low latency might also be required to shrink the sim-to-real gap: delayed sensing and actuation can degrade policy performance, requiring techniques to model or compensate for too-large latency~\citep{tan2018sim,schuitema2010control}.

One way to reduce latency is to move the policy to a faster compute platform, but this typically raises \emph{energy consumption}. That trade-off is especially problematic on battery-powered systems, such as nano-/micro-UAVs, which operate under power envelopes of a few watts or less~\citep{palossi2021fully}, as reduced energy draw directly extends mission time before recharge.  
For comparison, high-end GPUs routinely draw hundreds of watts, \eg up to \(700\,\mathrm{W}\) for one NVIDIA H100~\citep{nvidiaH100}.

\emph{Specialized embedded hardware}, notably field-programmable gate arrays (FPGAs), are a promising way to achieve low latency at low power~\citep{wan2021survey, finn}. However, these gains come with constraints: limited on-chip memory and the high cost of floating-point (FP32) arithmetic push deployments toward low-precision formats \citep{rati2025adaptive,finn}.
Overall, there is no universal deployment recipe for neural network policies: latency, energy, memory footprint, and numerical precision requirements vary by task and platform. In this work, we concentrate on \emph{continuous-control} neural network policies learned with RL in simulation, with the final inference target architecture being FPGAs. 

As FP32 operations are costly in hardware, we restrict ourselves to \emph{integer-only} operations, matching the computational abilities of FPGAs much better. Hence, standard FP32 networks cannot be deployed \emph{as is}. A straightforward path is {post-training quantization} (PTQ), which converts a pretrained full-precision model to an integer-only one using calibration data and no (or minimal) retraining~\citep{gholami2022survey}. 
However, PTQ often underperforms approaches that train with quantization in the loop: \emph{quantization-aware training} (QAT). %
Additionally, PTQ often performs worse on smaller models, which are investigated in this work~\citep{jacob2018quantization}. Furthermore, \citet{chandrinos2024quantization} report QAT outperforming PTQ in multi-agent few-bit quantization settings.

Therefore, we adopt QAT for continuous-control policies, training networks directly in the quantized regime. QAT explicitly accounts for limited numerical precision and the non-differentiability of quantization during training (via straight-through estimators), yielding models trainable with gradient descent and compatible with integer-only deployment.

\paragraph{Contributions.} We demonstrate that QAT can be successfully used for learning continuous-control policies, quantifying how far bitwidths and layer sizes can be reduced while preserving control quality relative to floating-point on MuJoCo tasks. We also examine the impact of quantization at the input interface. To probe robustness, we compare QAT and FP32 policies under input-noise injection. Finally, we synthesize trained policies to a representative FPGA using FINN~\citep{blott2018finn}, demonstrating better resource-latency trade-offs than a strong, already-quantized reference.

\section{Quantized Reinforcement Learning for Integer Hardware}
In this section, we outline our pipeline for training and deploying integer-only controllers: RL setup, QAT of the policy, and FPGA synthesis.

\subsection{Deep reinforcement learning for continuous control} 

In reinforcement learning, an agent interacts with its environment and, 
through trial and error, learns to maximize cumulative reward \citep{barto2021reinforcement}. This interaction is formalized as a Markov Decision Process (MDP) specified by the tuple $(\mathcal{S},\mathcal{A},P,R,\gamma)$, where $\mathcal{S}$ is the state space, $\mathcal{A}$ the action space, $P(s' \!\mid\! s,a)$ the transition dynamics, $R(s,a)$ the immediate reward, and $\gamma\!\in\!(0,1]$ the discount factor. At each time-step $t$, the agent follows a policy $\pi$ selecting an $a_t\!\in\!\mathcal{A}$ given the current state $s_t\!\in\!\mathcal{S}$; the environment then samples the next state $s_{t+1}\!\sim\!P(\cdot\!\mid\!s_t,a_t)$ and returns a reward $r_t\!=\!R(s_t,a_t)$. An \emph{episode} is the resulting sequence $(s_0,a_0,r_0,s_1,\ldots)$, and the learning  objective is to maximize the expected discounted return
\(
G \;=\; \sum_{t=0}^{N-1} \gamma^{t}\, r_t .
\)

In this work, we consider the setting where both $\mathcal{S}$ and  $\mathcal{A}$ are continuous and multi-dimensional. A variety of algorithms have been proposed to solve for maximizing the discounted return in this setting. We focus on off-policy, actor-critic methods, specifically Deep Deterministic Policy Gradient (DDPG)~\citep{lillicrap2015continuous} and Soft Actor-Critic (SAC)~\citep{haarnoja2018soft}. 
DDPG uses a deterministic \emph{policy}, \(\pi_\theta:\mathcal{S}\to\mathcal{A}\), 
that proposes actions for a given state, together with a \emph{critic}, \(Q_\phi:\mathcal{S} \times \mathcal{A} \to \mathbb{R}\), that estimates expected returns.
At training time, SAC employs a {stochastic} Gaussian policy 
\(\pi_\theta(a\!\mid\!s)=\mathcal{N}\!\big(\mu_\theta(s),\mathrm{diag}(\sigma_\theta^2(s))\big)\) together with two critics.
At deployment-time, SAC uses the deterministic maximum-likelihood action.
In both cases, the parameters \(\theta\) and \(\phi\) are neural-network 
weights that are optimized end-to-end via gradient-based updates based on 
a replay buffer of transitions.

\subsection{Quantization-aware training}\label{sec:QAT}

We instantiate all neural networks, \ie the policy and the critics, as standard full-precision (FP32) neural networks with ReLU non-linearities and a final hyperbolic tangent activation function that bounds action values to \([-1,1]\), following the implementation in CleanRL~\citep{huang2022cleanrl}. 
Because our goal is integer \emph{inference}, and the critics networks can be discarded after training, we must only ensure that the policy network follows integer arithmetic. For SAC, the $\sigma_\theta$ branch is not needed at deployment; we therefore implement it as a {separate} FP32 subnetwork (one hidden layer with 64 units) used only during training.

To ensure the policy $\pi_\theta$ is compatible with integer arithmetic deployment, we impose \emph{Quantize/De-Quantize} (\QDQ) steps~\citep{jacob2018quantization}, at the input, for all weights, after every ReLU, and before the final output. For a positive scale $s$  and a FP32 input $x$, this is defined as
\begin{equation}\label{eq:qdq}
\QDQ_b(x;s)\;=\;\frac{s}{q_s}\,\Q_b(x;s),\qquad
\Q_b(x;s)\;=\;\operatorname{clip}\big(\operatorname{round}\!\big(\tfrac{x}{s} q_s\big),\,q_{\min},\,q_{\max}\big),
\end{equation}
where $\operatorname{round}$ rounds a value to its nearest integer. 
The integer bounds $(q_{\min},q_{\max})$ are fixed by the bitwidth parameter $b$ and the intended signedness of the operation: for the inputs, the network weights, and the final output we use \emph{signed quantization}: $[q_{\min},q_{\max}]=[-2^{b-1},\,2^{b-1}-1]$. For all intermediate layers, a ReLU activation precedes the \QDQ step, so negative values cannot occur, and we use \emph{unsigned quantization} to fully utilize the available bitwidth: $[q_{\min},q_{\max}]=[0,\,2^{b}-1]$. The to-integer scaling factor $q_s$ is defined as $q_s=\max(|q_{\min}|,|q_{\max}|)$. \QDQ projects the FP32 value onto the integer lattice (via scaling, rounding, and clipping) and then de-quantizes it back to FP32, allowing us to retain a standard FP32 training loop while enforcing integer arithmetic at deployment.

Because rounding is piecewise constant and therefore non-differentiable, we use a \emph{straight-through estimator} (STE) to backpropagate through \QDQ: during the backward pass, we treat the quantizer as the identity for gradients~\citep{bengio2013estimating, jacob2018quantization}.

In this work, we initialize the activation scales during a 300-step warm-up via an exponential moving high percentile of the input statistics and, after warm-up, we learned them by gradient updates. Weight scales are not learned: for each weight matrix, the scale is fixed to its absolute maximum.

\subsection{Integer-only deployment}\label{sec:deployment}
At deployment time, all quantization scales are fixed. For each weight tensor \(w\) with associated scale \(s_w\), we straightforwardly obtain its integer representation \(\tilde{w}=\Q_b(w;s_w)\). The input state \(x_0\) is quantized on the fly using the floating-point input scale \(s_{x}\), \ie , \(\tilde{x}_0=\Q_b(x_0;s_{x})\), and thereafter the network runs integer-only. 
Each layer performs integer matrix-vector products with sufficiently wide accumulators to avoid overflow; after accumulation ReLU is applied, and the result is \emph{requantized} to the target bitwidth for the next layer, \ie, it is scaled according to the predetermined scale of the next layer and clipped to \([q_{\min},q_{\max}]\). 
In our FINN-based deployment, this requantization is implemented based on stored thresholds~\citep{blott2018finn}, thus avoiding any FP32 operations. 
At the final layer, an analogous requantization takes place with a mapping to \([-1,1]\) via a hyperbolic tangent (implemented as a lookup), yielding the action value for the environment.

Ultimately, the only floating-point operation involved is the {initial} quantization of the input state. 
We leave this unmodified, because for real-world systems the needed processing
step depends on the sensors used. 
For example, sensor readings typically arrive as integers (\eg, from analog-digital converters), which can be scaled via integer arithmetic or lookup tables. %

To deploy the quantized policy network on a hardware platform, it has 
to be \emph{synthesized}. For this, we use the FINN dataflow compiler~\citep{finn,blott2018finn}, 
generating a \emph{streaming dataflow}. %
Layers are connected by FIFO streams, and both weights and activations are kept on chip, eliminating external DRAM traffic and thereby reducing latency and energy consumption. 
Each layer exposes two parallelism variables: the number of \emph{processing elements} (PEs; parallelism along the matrix {rows} / output features) and the \emph{SIMD} width (parallel inputs consumed per cycle, \ie, columns). Increasing either reduces per-layer cycle count and raises throughput, at the cost of additional resources. FINN can also target a user-specified throughput and automatically choose PE and SIMD count to meet it.

\section{Experiments}
\label{sec:experiments}

\begin{figure}
\begin{tikzpicture}
\begin{groupplot}[
  group style={group size=2 by 5, horizontal sep=1.1cm, vertical sep=0.9cm},
  myaxis,
  legend to name=globalLegend, legend columns=5,
]

\nextgroupplot[title={\textbf{SAC}},ymin=0, ymax=7000,ylabel=\shortstack{\textbf{Humanoid-v4}\\Reward}]
\envpanel{data/sac_new}{Humanoid-v4}

\nextgroupplot[title={\textbf{DDPG}},ymin=0, ymax=7000,yticklabels=\empty, ylabel={},]
\envpanel{data/ddpg_mixed}{Humanoid-v4}

\nextgroupplot[title={},ymin=0, ymax=7000,ylabel=\shortstack{\textbf{Walker2d-v4}\\Reward}]
\envpanel{data/sac_new}{Walker2d-v4}

\nextgroupplot[title={},ymin=0, ymax=7000,yticklabels=\empty, ylabel={},]
\envpanel{data/ddpg_mixed}{Walker2d-v4}

\nextgroupplot[title={}, ymin=0, ymax=7000,ylabel=\shortstack{\textbf{Ant-v4}\\Reward}]
\envpanel{data/sac_new}{Ant-v4}

\nextgroupplot[title={},ymin=0, ymax=7000,yticklabels=\empty, ylabel={}]
\envpanel{data/ddpg_mixed}{Ant-v4}

\nextgroupplot[title={},ymin=0, ymax=12000,ylabel=\shortstack{\textbf{HalfCheetah-v4}\\Reward}]
\envpanel{data/sac_new}{HalfCheetah-v4}

\nextgroupplot[title={},ymin=0, ymax=12000,yticklabels=\empty, ylabel={},]
\envpanel{data/ddpg_mixed}{HalfCheetah-v4}

\nextgroupplot[title={ },ymin=0, ymax=4000,ylabel=\shortstack{\textbf{Hopper-v4}\\Reward}]
\envpanel{data/sac_new}{Hopper-v4}

\nextgroupplot[title={},ymin=0, ymax=4000,yticklabels=\empty,ylabel={}]
\envpanel{data/ddpg_mixed}{Hopper-v4}

\end{groupplot}
\node at (6.2,-14.5) {\pgfplotslegendfromname{globalLegend}};
\end{tikzpicture}
\caption{{Reward \vs bitwidth} for full-precision (FP32) baselines (shaded region indicates one standard deviation) as well as four variants of network quantization: \emph{all}: all network operations are quantized to indicated bitwidths; \emph{input}/\emph{output}: the quantization of only the inputs/outputs are varied; \emph{core}: the quantization of weights and internal activations are varied. In the latter three cases, all other components are left at 8-bit precision. We achieve FP32-parity with SAC and DDPG across most quantization scopes and environments. See main text for further discussion of the curves.}
\label{fig:main results}
\end{figure}
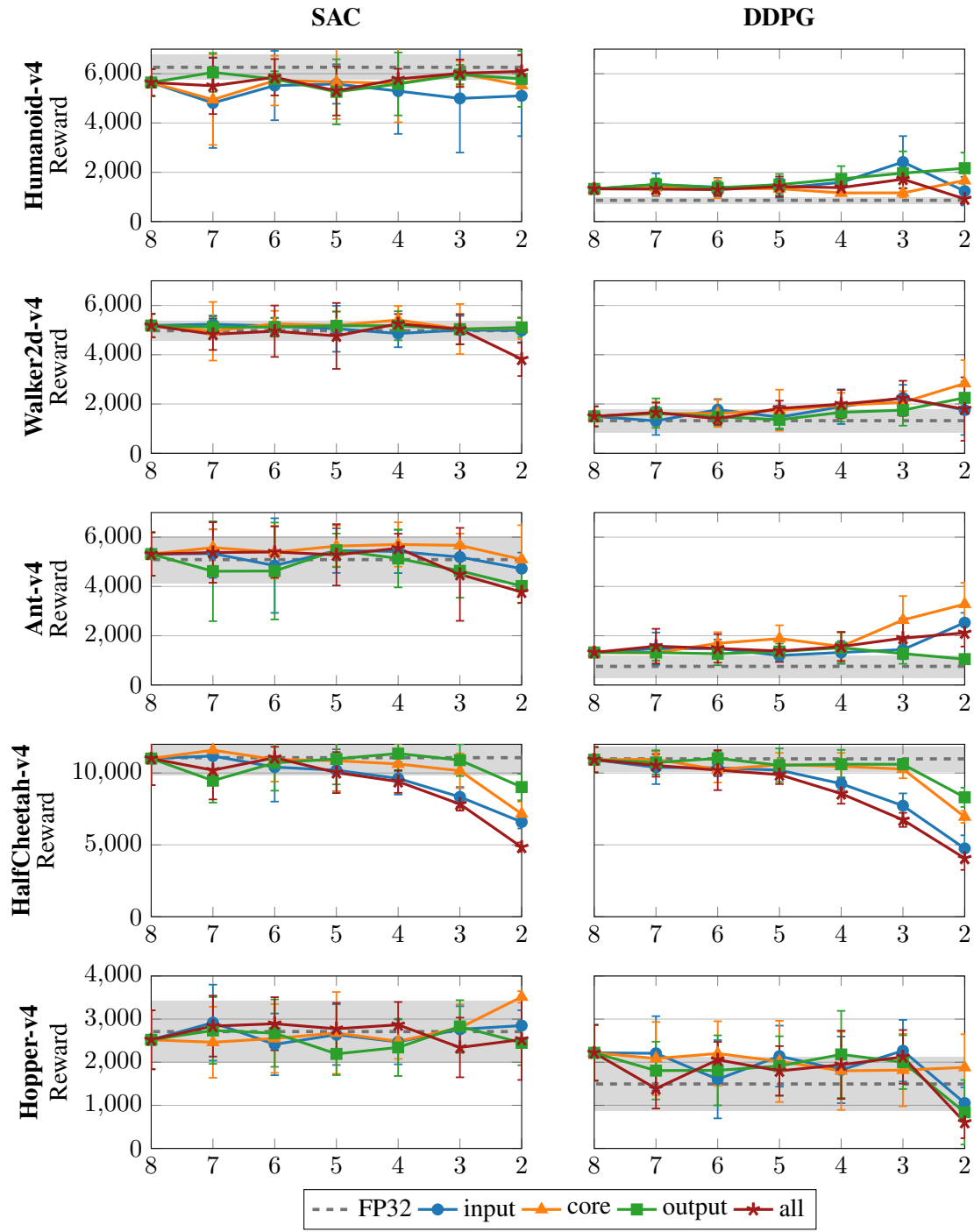

We evaluate QAT on five MuJoCo tasks (Humanoid-, Walker2d-, Ant-, HalfCheetah-, Hopper-v4) using SAC and DDPG. %
Our goal is to obtain \emph{very small}, hardware-friendly policies that match FP32 returns. We structure the evaluation into three parts: (i) \emph{bitwidth sensitivity}, where we vary which components of the policy are quantized to investigate where precision matters; (ii) \emph{model selection}, a three-step rule that first fixes the smallest FP32 matching configuration for internal activations and weight bitwidth, then the hidden layer width, and last the input activation precision; and (iii) \emph{robustness}, where we perform noise-injection tests on input states. Finally, we synthesize the selected policies to an Artix-7 FPGA with FINN v0.10.1, targeting a fixed \(100\,\mathrm{MHz}\) clock and reporting latency, throughput, power usage, and resources utilized.

All training runs were performed with 1M environment steps; for evaluation, we average over 1{,}000 (10 where stated) deterministic policy rollouts per trained model, and report mean and standard deviation across 10 training seeds. All reported rewards are undiscounted. \\
\textbf{Baseline (FP32)} As a baseline, we train full-precision policies using the CleanRL~\citep{huang2022cleanrl} implementations of SAC and DDPG, utilizing corresponding default hyperparameters and model sizes (see Appendix \ref{app:hyperparameters}). In addition, we perform running normalization of the input state space (per-dimension, frozen at evaluation) {for SAC, but not DDPG,} as we find it to perform better. %
\\
\textbf{Quantized Models} We swap the policy network for a QAT variant implemented with Brevitas v0.12.0~\citep{brevitas}, using the same hyperparameters as the FP32 baselines. {In this case, we perform running normalization for both DDPG and SAC.} An ablation on normalization can be found in Appendix \ref{app:normalization}.

\subsection{Bitwidth Sensitivity}
To determine where quantization matters, we sweep bitwidth for four quantization scopes, always using QAT for all weights, activations, and biases but fixing non-swept parts to 8-bit. We investigate four scopes. \emph{all}: all network components (input state activation, output, weights, and other activations) are quantized to varying bitwidths; \emph{input}: only the bitwidth used to represent the input state is varied; \emph{output}: only the final pre-hyperbolic tangent activation requantization is varied; \emph{core}: all internal weights and activation values are quantized. %
Figure~\ref{fig:main results} reports returns for each scope and bitwidth; shaded bands show mean plus/minus one standard deviation across 10 models per data-point. We consider a quantized model to \emph{match} FP32 returns if its mean return lies within the FP32 band. %

\paragraph{SAC} For SAC, we find that our quantized models generally match the FP32 baseline down to at least 3 bits, even in the \emph{all} setting. The only major exception is HalfCheetah, where we find that starting at 5-bits reward decreases. This is consistent with a decreasing reward in the \emph{input} setting, indicating that input quantization alone can have a major impact on reward. However, we also find that for HalfCheetah the \emph{core} setting for 2-bit results in a reward drop; hinting at model capacity-limited return, which is further supported by layer-width reduction results in Appendix~\ref{app:ms}. For Humanoid, we observe that the variance between policies is often higher across bitwidths than the FP32 policy. However, both the \emph{core} and \emph{all} settings clearly match the FP32 baseline at 3-bits, with similar variances. Last, we observe that output quantization only appears to matter at the extreme of 2-bits. 
\paragraph{DDPG} {With DDPG, we observe lower overall rewards, though the quantized versions generally match the FP32 baselines. HalfCheetah shows a performance drop similar to the one seen with SAC. Notably, for the Ant task, lower bit-width quantization appears to improve rewards. We hypothesize that this behavior relates to findings that RL performance does not always improve monotonically with model capacity~\citep{bjorck2021towards}.  } %

For both algorithms, the \emph{output} quantization has the least influence on performance. However, as the output quantization also has only a minor influence on the size of the synthesized hardware model, when compared to other parameters, we leave it at 8-bit for the remainder of this work.

In our environments, if FP32 parity cannot be reached in the \emph{all} setting, the input bitwidth can generally be considered the bottleneck, as we observe the \emph{core} and \emph{output} setting consistently outperforming the \emph{input} setting. {We attribute this to the fact that high quantization error in the first layer prevents the network from clearly differentiating between similar sensor states, causing a loss of information important for fine control. In contrast, once the signal passes the first layer, the information is distributed across a latent space with more redundancy. This makes the deeper layers more resilient to lower bit-widths, as the network is no longer relying on the high-precision resolution of a single input feature. Similar results of the importance of the first layer have also been found, for instance, regarding sparsity in convolutional neural networks \citep{han2015learning}}. As SAC outperforms DDPG in our setting, we use it in the remainder of this work.

\subsection{Model Selection}
\begin{figure*}[t]
\centering

\begin{tikzpicture}
\begin{groupplot}[
  group style={
    group size=3 by 1,
    horizontal sep=1.1cm,
    vertical sep=1.3cm
  },
  timeaxis,
  legend to name=globalLegend2, legend columns=4,
]
  \nextgroupplot[title={Humanoid-v4}, ymin=0, ymax=7000, ylabel={Reward}]
    \bitpanel{data/curves_new}{Humanoid-v4}{3}
      \nextgroupplot[title={Walker2d-v4}, ymin=0, ymax=7000]
    \bitpanel{data/curves_new}{Walker2d-v4}{2}
  \nextgroupplot[title={Ant-v4}, ymin=0, ymax=7000, ylabel={}]
    \bitpanel{data/curves_new}{Ant-v4}{2}

\end{groupplot}
\end{tikzpicture}

\makebox[\textwidth][c]{%
\begin{tikzpicture}
\begin{groupplot}[
  group style={
    group size=2 by 1,
    horizontal sep=1.1cm,
    vertical sep=1.3cm
  },
  timeaxis,
  legend columns=4,
  legend to name=globalLegend2
]
  \nextgroupplot[title={HalfCheetah-v4}, ymin=0, ymax=12000,  ylabel={Reward}]
    \bitpanel{data/curves_new}{HalfCheetah-v4}{3}

  \nextgroupplot[title={Hopper-v4}, ymin=0, ymax=4000]
    \bitpanel{data/curves_new}{Hopper-v4}{2}
    
\end{groupplot}
\node at (4.0,-1.5) {\pgfplotslegendfromname{globalLegend2}};
\end{tikzpicture}%
}
\caption{Evaluation reward across training time steps for our environments with SAC. Shaded bands show standard deviation over trained models. Overall, the selected quantized models show comparable convergence behavior to the floating-point baseline.}
\label{fig:training_steps}
\end{figure*}
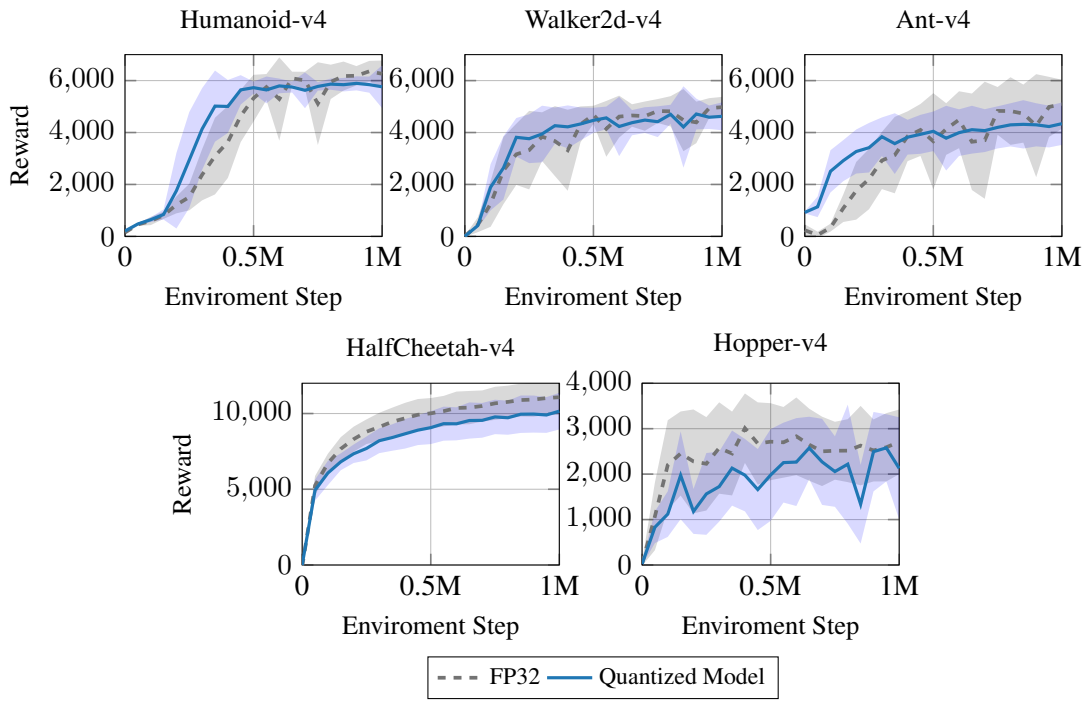

\label{sec:selection}

Guided by the sensitivity results, we select SAC configurations under the previous FP32-parity criterion (mean within the FP32 mean plus/minus one standard deviation). 
First, we choose the smallest core quantization precision, a single bitwidth  ($b_{\mathrm{core}}$) shared by hidden-layer weights and internal activations, that matches FP32 with input/output fixed at 8-bit (Figure~\ref{fig:main results}). We prioritize $b_{\mathrm{core}}$ because it strongly affects hardware cost (e.g., requantization resources scale exponentially with activation bitwidth in FINN~\citep{blott2018finn}). Next, holding that precision, we sweep the hidden layer width \(h\) over \(\{256,128,64,32,16\}\) and pick the smallest that still matches FP32. Finally, with $b_{\mathrm{core}}$ and \(h\) fixed, we sweep input state quantization bits ($b_{\mathrm{in}}$). Figures showing sweeps for model selection are available in Appendix \ref{app:ms}.

\begin{wraptable}[12]{r}{6.0cm}
\vspace{-1\baselineskip}
\centering
\setlength{\tabcolsep}{3.5pt}   %
\renewcommand{\arraystretch}{0.9}

\begin{tabular}{l l l l}
\toprule
Environment & {$h$} & {$b_{\mathrm{core}}$} & {$b_{\mathrm{in}}$} \\
\midrule
Humanoid   & 16  & 3 & 4 \\
Walker2d   & 128 & 2 & 3 \\
Ant        & 64  & 2 & 3 \\
HalfCheetah& 256 & 3 & 8 \\
Hopper     & 16  & 2 & 6 \\
\bottomrule
\end{tabular}

\caption{Selected SAC policies: \emph{hidden layer width} $h$; \emph{core bitwidth} $b_{\mathrm{core}}$ (weights and activations); \emph{input bitwidth} $b_{\mathrm{in}}$.}
\label{tab:final-config}
\end{wraptable}

Table~\ref{tab:final-config} reports the resulting models. The selected models evaluation reward over training environment steps is shown in Figure~\ref{fig:training_steps}, with 10 rollouts performed per evaluation step, per model. With our staged selection, FP32 parity is always achieved with \emph{3-bit cores}, often even 2 bits suffice. The tolerable hidden-width reduction and observation quantization are environment-dependent. Furthermore, the acceptable {input activation} bitwidth shrinks as core precision and hidden layer width is reduced: compare Figure~\ref{fig:main results} (input sweep with core fixed) to Table~\ref{tab:final-config} (after fixing the minimal FP32-parity $b_{\mathrm{core}}$ and $h$), where the attainable input precision is generally lower.

\subsection{Noise injection}

\begin{figure}
    \centering

 \begin{tikzpicture}    
    \begin{groupplot}[
      group style={
        group size=5 by 1,          %
        horizontal sep=1.1cm,
        vertical sep=1.3cm,
      },
      stdaxis,                     %
      legend columns=4,
      legend to name=globalLegend5,
      xmin=0, xmax=5,
      xtick={0,1,2,3,4,5},
      xticklabels={.0,.1,.2,.3,.4,.5},
      scaled y ticks=false,
      ymin=0,
      xticklabel style={font=\scriptsize}
    ]
    \nextgroupplot[title={Humanoid-v4}]
      \noisepanel{data/noise}{Humanoid-v4}
              \nextgroupplot[title={Walker2d-v4}, ylabel={}]
      \noisepanel{data/noise}{Walker2d-v4}
          \nextgroupplot[title={Ant-v4}, ylabel={}]
      \noisepanel{data/noise}{Ant-v4}
              \nextgroupplot[title={HalfCheetah-v4},ylabel={}]
\noisepanel{data/noise}{HalfCheetah-v4}

    \nextgroupplot[title={Hopper-v4}, ylabel={}]
      \noisepanel{data/noise}{Hopper-v4}

        \showlegendtrue      

      \showlegendfalse
    
    \end{groupplot}
     \node at ($(group c1r1.north)!0.5!(group c5r1.north)$) [yshift=-30mm, xshift=0mm]
    {\pgfplotslegendfromname{globalLegend5}};
 \end{tikzpicture}
 \caption{Robustness to observation input noise. Reward \vs noise level, $\sigma$, for floating-point and selected QAT policies on MuJoCo tasks. Shaded bands show standard deviation over trained models. The quantized, selected model performs better, or on par, with the FP32 baseline under injection.}
 \label{fig:noise}
\end{figure}
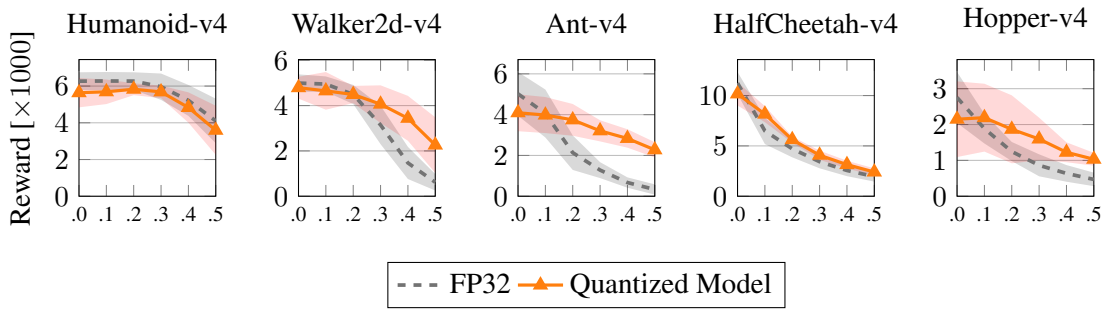

We evaluate robustness to input state perturbations by adding i.i.d.\ Gaussian noise to the \emph{normalized} state at inference: 
$\hat{s}=\mathrm{norm}(s)+\epsilon$, $\epsilon\sim\mathcal{N}(0,\sigma^2)$ with $\sigma\in\{0.0,0.1,0.2,0.3,0.4,0.5\}$. 
We compare the \emph{selected} quantized model (Table~\ref{tab:final-config}) with the FP32 baseline.
For each $\sigma$ and task we average returns over $10$ models ($10$ episodes each) and report the standard deviation of model means. 
Figure~\ref{fig:noise} shows return vs.\ $\sigma$. Across tasks, quantized policies match or exceed FP32 at higher noise levels for the quantized models. {We hypothesize this to be due to training time discretization of the state space. While very similar states are often mapped to the same quantization bins, which filters out small noise, equally small noise can lead to changes in the utilized bins. To achieve high returns despite these discrete jumps for very similar states, the policy is forced to remain stable across adjacent bins. In contrast, the FP32 baseline may react to even small perturbations.}

\subsection{FPGA Synthesis}

\begin{wraptable}[10]{r}{6cm}
\vspace{-\baselineskip}
  \begin{tabular}{l r}
    \toprule
    Resource & Quantity \\
    \midrule
    Lookup-Tables (LUTs) &10{,}400 \\
    Flip-flops (FFs) & 20{,}800 \\
    Block RAM (36\,Kb) & 25 \\
    Digital Signal Processing & 45 \\
     Units (DSPs) & \\
    \bottomrule
  \end{tabular}
    \caption{XC7A15T-FGG484\(-1\) device resources.}
      \label{tab:artix}
\end{wraptable}

We synthesize policies from Table~\ref{tab:final-config} for an Artix-7 {XC7A15T} (speed grade \(-1\)); available device resources are listed in Table~\ref{tab:artix}. {Note that this is a low-end FPGA, having a limited number of resources. Larger FPGAs could improve throughput and latency.} For all networks, we only investigate the performance and hardware requirements for the integer portion; hence, initial floating-point quantization and final hyperbolic tangent mapping are not considered. Networks are exported with \textsc{FINN} and built out-of-context (OOC) as IP cores without a board wrapper. All action dimensions are padded to multiples of 32 for compatibility with FINN. %

Synthesis follows a throughput-driven procedure: for each policy, we sweep target throughputs in powers of 10, while FINN selects folding factors, SIMD width and the number of PEs, to approximately hit the target at a fixed \(100\,\mathrm{MHz}\) clock. We retain the highest target whose build completes and meets timing. The resulting resources, latency and estimated power are summarized in Table~\ref{tab:hardware}.

8-bit models exceed {XC7A15T} capacity, because FINN’s integer-only requantization memory cost grows exponentially with bitwidth~\citep{blott2018finn}. We therefore use a reference that preserves the original architecture (width 256) with a 4-bit \emph{core} (I/O at 8 bit) and the prior folding selection. Except for Humanoid, these designs use similar resources as padding yields matched widths; Humanoid differs due to its larger input dimension. Latency and energy per action improve for all selected models relative to this reference except HalfCheetah, where our selection (256, 3-bit) mainly reduces BRAM usage compared to the 4-bit reference. Walker2d gains a $\sim\!1.5\times$ speedup while remaining wide (128), but overall resource and power needs drop substantially. Ant and Humanoid exhibit order-of-magnitude latency gains (\(10\!\times\)--\(100\!\times\)), and Hopper reaches a 21-cycle latency, corresponding to a \(\sim\!1000\!\times\) improvement. Note that a 4-bit reference is already a strong baseline; prior work \citep{krishnan2019quarl} reported collapse on Walker2d at 4-bit quantization.

\sisetup{
  scientific-notation = true,   %
  exponent-product   = \times,  %
  table-number-alignment = center
}

\begin{table*}[t]
  \centering
  \scalebox{0.9}{\begin{tabular}{
    l                         %
    l                         %
    l                         %
    l                         %
    l                         %
    l                         %
    S[detect-weight = true,   %
      round-mode = places,
      round-precision = 1,
      scientific-notation = fixed,
      fixed-exponent = 3,
      exponent-to-prefix = true,]
    l       %
    S[table-format=1.3e2]     %
    S[table-format=1.2e1]     %
  }
    \toprule
    & \multicolumn{1}{c}{Environment} &
      \multicolumn{1}{c}{LUTs} &
      \multicolumn{1}{c}{FFs} &
      \multicolumn{1}{c}{BRAM} &
      \multicolumn{1}{c}{DSP} &
      \multicolumn{1}{c}{Latency} &
      \multicolumn{1}{c}{P [\si{\watt}]} &
      \multicolumn{1}{c}{TP} &
      \multicolumn{1}{c}{E.p.A. [\si{\joule}]}\\
    \midrule

    \multirow{5}{*}{\rotatebox{90}{\shortstack{Selected \\ Model}}}
     & Humanoid    & \numk{2256} & \numk{3118} & 1.5 & 45 & \phantom{$0$}\ns{15360}  & 0.33 & \num{65104}    & \num{5.068812976161219e-6} \\
      & Walker2D    & \numk{1858} & \numk{1613} & 2   & 4  & \ns{162230} & 0.17 & \num{6164}     & \num{2.757949383517197e-5} \\
      & Ant         & \numk{2699} & \numk{4497} & 3   & 45 & \phantom{$00$}\ns{2290}   & 0.39 & \num{436681}   & \num{8.93e-7} \\
      & HalfCheetah & \numk{4336} & \numk{4565} & 15  & 11 & \ns{243230} & 0.33 & \num{4111}     & \num{8.02e-5} \\
    & Hopper      & \numk{2377} & \numk{1994} & 0   & 45 & \phantom{$00$}\ns{210}    & 0.31 & \num{4761904}& \num{6.510001041600167e-8} \\
    \midrule\midrule
    \multirow{5}{*}{\rotatebox{90}{\shortstack{Reference \\(8-4-8)}}}
    & Humanoid    & \numk{8426} & \numk{9545} & 25  & 22 & \ns{243270} & 0.58 & \num{4110}     & \num{1.411192214111922e-4} \\
      & Walker2D    & \numk{5754} & \numk{5740} & 25  & 11 & \ns{243250} & 0.40 & \num{4110}     & \num{9.732360097323601e-5} \\
      & Ant         & \numk{4648} & \numk{5916} & 25  & 11 & \ns{243250} & 0.37 & \num{4110}     & \num{9.002433090024331e-5} \\
      & HalfCheetah & \numk{5285} & \numk{5730} & 25  & 11 & \ns{243250} & 0.38 & \num{4110}     & \num{9.245742092457421e-5} \\

      & Hopper      & \numk{4435} & \numk{5404} & 25  & 11 & \ns{243250} & 0.35 & \num{4110}     & \num{8.515815085158151e-5} \\
    \bottomrule
  \end{tabular}}
  \caption{Post-synthesis resource utilization, end-to-end latency, power estimated by the Xilinx Power Estimator (P) in Watts, peak throughput (TP) in actions per second, and energy per action (E.p.A.) in Joule on a Artix-7 {XC7A15T}\(-1\) at \(100\,\mathrm{MHz}\). }
  \label{tab:hardware}
\end{table*}

\section{Related Work}

\paragraph{Quantized training for RL.} The most closely related works are \citet{krishnan2019quarl, lu2024impact} and \citet{ivanov2025neural}, who also study QAT for MuJoCo tasks. 
Of these, only \citet{krishnan2019quarl} investigates QAT for low bitwidths in MuJoCo, as we do here.  
Specifically, they study both post-training quantization (PTQ) in a distributed actor-learner setup, and QAT on Atari and some MuJoCo tasks for different RL methods, including DDPG but not SAC. 
Unfortunately, their public implementation does not cover the low-bitwidth experiments, and key QAT details are missing from the manuscript.
Known setting differences include a much larger budget of environment steps (10M steps), where QAT is enabled only after 5M steps, whereas we train for 1M steps with QAT from scratch. Furthermore, the environment versions differ.
Therefore, only a qualitative comparison with their results makes sense: 
\citet{krishnan2019quarl} report QAT experiments for HalfCheetah and Walker2d with DDPG. For both tasks, they report substantial return degradation starting at 4 to 5 bits. In our work, the \emph{core} setting for these tasks shows reward degradation only at 2 bits, or none, respectively (Figure \ref{fig:main results}, DDPG). Note that the baseline reward they obtain with Walker2d is comparable to ours, whereas our baseline obtains much higher returns for HalfCheetah.

\citet{lu2024impact} find that 8-bit QAT under their protocol performs worse than FP32 networks on MuJoCo tasks. \citet{ivanov2025neural} pretrain in full precision and then fine-tune with QAT for 8-bit \emph{weights} while keeping \emph{activations} in FP32, thereby avoiding activation quantization error but also limiting end-to-end efficiency gains. 
\citet{chandrinos2024quantization} study both PTQ and QAT in a multi-agent environment with discrete actions. They find that QAT substantially outperforms PTQ.

\citet{gil2021quantization} study PTQ for continuous control by quantizing and pruning pretrained policies. They find that 4-bit weights %
largely preserve return on a real inverted pendulum setup, whereas 3-bit and below resulted in a substantial loss of reward.

\paragraph{Pruning for RL.} Pruning---removing parameters or connections---can reduce multiply-accumulate (MAC) operations and memory traffic at inference. In reinforcement learning, very high levels of {unstructured} sparsity (often $>95\%$) have been reported without return loss across diverse settings \citep{ivanov2025neural,tan2022rlx2,arnob2024efficient,graesser2022state}, and scaling studies suggest that larger, and pruned, policies can even improve reward scaling \citep{ma2025network}. Pruning has also been studied in combination with QAT~\citep{lu2024impact, ivanov2025neural}. 
However, unstructured sparsity is generally difficult to exploit in hardware, and we therefore do not consider it in our work. 
Nevertheless, these works show that standard RL networks are typically severely overparameterized, inspiring our architecture search. 

\paragraph{Related Directions.}
At the quantization extreme, work on binarized policies targets discrete action spaces \citep{kadokawa2021binarized,lazarus2022deepbinaryreinforcementlearning,valencia2019using}, with some studies binarizing only parts of the network~\citep{chevtchenko2021combining}. Additionally, several works discretize only parts of the RL pipeline for practical reasons: discretizing the continuous action space to change the algorithmic regime~\citep{dadashi2021continuous}, or quantizing inputs to reduce replay-buffer memory with negligible return loss~\citep{grossman2022just}. Last, performing training in half-precision has been explored \citep{bjork2021lowprecisionrl}. These target different bottlenecks than end-to-end integer inference.

\section{Discussion \& Future Work}
In this work, we show how to use quantization-aware training to obtain deployment-ready, integer-only policy networks that match the returns of full-precision networks on MuJoCo tasks.
A key finding of our work is that network weights and internal activations can be quantized quite aggressively (a bitwidth of 2 or 3 suffices), allowing an implementation on low-resource FPGA hardware with sub-millisecond inference latency and consuming microjoules per action inference.
Furthermore, \emph{input} quantization should be treated separately, because it can have a major influence on the policy quality.
We also find that quantized models are often more robust under input noise, 
presumably because the noisier gradient updates during training act as implicit regularization. 
We therefore hypothesize that QAT-trained policies may be more robust in the real world under sensor noise.

Despite the promising results, a number of limitations and open questions remain.
First, our study reflects the situation of training and deployment in simulation only. Clearly, for real-world systems, additional complications can emerge. 
For example, aspects such as safety and security need to be taken into account besides the expected reward that we use in our analysis to judge policy quality. 
Also, real-world noise can have different and more challenging characteristics than the analytically injected noise in our experiments. 

For future work, it would be interesting to move beyond classical MuJoCo tasks and investigate additional scenarios, such as vision-based policies for mobile robots {or sparse rewards}.
An additional direction is evaluation on different hardware targets (other FPGAs and non-FPGA platforms) and on physical systems to capture sensor/actuator overheads and measured power.

\acks{
This work was partially supported by the Austrian Science Fund (FWF) [10.55776/COE12]. This work was supported in part by the ERC-2020-AdG 101020093 (VAMOS).}
The research was supported by the Scientific Service Units (SSU) 
of ISTA through resources provided by Scientific Computing (SciComp).

\bibliography{main}

@string{NeurIPS = "Conference on Neural Information Processing Systems (NeurIPS)"}

@string{ICLR = "International Conference on Learning Representations (ICLR)"}

@string{ICML = "International Conference on Machine Learning (ICML)"}

@string{ICRA = "IEEE International Conference on Robotics and Automation (ICRA)"}

@string{CVPR = "IEEE/CVF Conference on Computer Vision and Pattern Recognition (CVPR)"}

@string{RSS = "Robotics Science and Systems (RSS)"}

@software{brevitas,
  author       = {Franco, Giuseppe and Pappalardo, Alessandro and Fraser, Nicholas J},
  title        = {Xilinx/brevitas},
  year         = {2025},
  publisher    = {Zenodo},
  doi          = {10.5281/zenodo.3333552},
  version = {???},
  url          = {https://doi.org/10.5281/zenodo.3333552}
}

@inproceedings{haarnoja2018soft,
  title={Soft actor-critic: Off-policy maximum entropy deep reinforcement learning with a stochastic actor},
  author={Haarnoja, Tuomas and Zhou, Aurick and Abbeel, Pieter and Levine, Sergey},
  booktitle=ICML,
  nopages={1861--1870},
  year={2018},
  noorganization={Pmlr}
}

@inproceedings{lillicrap2015continuous,
      title={Continuous control with deep reinforcement learning}, 
      author={Timothy P. Lillicrap and Jonathan J. Hunt and Alexander Pritzel and Nicolas Heess and Tom Erez and Yuval Tassa and David Silver and Daan Wierstra},  
      booktitle=ICLR,
      year={2016},
}

@article{kaufmann2023champion,
  title={Champion-level drone racing using deep reinforcement learning},
  author={Kaufmann, Elia and Bauersfeld, Leonard and Loquercio, Antonio and M{\"u}ller, Matthias and Koltun, Vladlen and Scaramuzza, Davide},
  journal={Nature},
  novolume={620},
  nonumber={7976},
  nopages={982--987},
  year={2023},
  nopublisher={Nature Publishing Group UK London}
}

@article{wan2021survey,
  title={A survey of {FPGA}-based robotic computing},
  author={Wan, Zishen and Yu, Bo and Li, Thomas Yuang and Tang, Jie and Zhu, Yuhao and Wang, Yu and Raychowdhury, Arijit and Liu, Shaoshan},
  journal={IEEE Circuits and Systems Magazine},
  novolume={21},
  nonumber={2},
  nopages={48--74},
  year={2021},
  nopublisher={IEEE}
}

@incollection{gholami2022survey,
  title={A survey of quantization methods for efficient neural network inference},
  author={Gholami, Amir and Kim, Sehoon and Dong, Zhen and Yao, Zhewei and Mahoney, Michael W and Keutzer, Kurt},
  booktitle={Low-power computer vision (LPCV)},
  nopages={291--326},
  year={2022},
  nopublisher={Chapman and Hall/CRC}
}

@inproceedings{chandrinos2024quantization,
  title={Quantization-Aware Training for Multi-Agent Reinforcement Learning},
  author={Chandrinos, Nikolaos and Amasialidis, Michalis and Kirtas, Manos and Tsampazis, Konstantinos and Passalis, Nikolaos and Tefas, Anastasios},
  booktitle={European Signal Processing Conference (EUSIPCO)},
  nopages={1891--1895},
  year={2024},
  noorganization={IEEE}
}

@article{kato2015open,
  title={An open approach to autonomous vehicles},
  author={Kato, Shinpei and Takeuchi, Eijiro and Ishiguro, Yoshio and Ninomiya, Yoshiki and Takeda, Kazuya and Hamada, Tsuyoshi},
  journal={IEEE Micro},
  novolume={35},
  nonumber={6},
  nopages={60--68},
  year={2015},
  nopublisher={IEEE}
}

@misc{nvidiaH100,
    author={{NVIDIA}},
    year={2025},
    title= {{NVIDIA H100 Data Sheet}},
    url= {{https://resources.nvidia.com/en-us-gpu-resources/h100-datasheet-24306}},
    note = {Accessed: 2025-11-06}
}

@article{palossi2021fully,
  title={Fully onboard {AI}-powered human-drone pose estimation on ultralow-power autonomous flying nano-{UAV}s},
  author={Palossi, Daniele and Zimmerman, Nicky and Burrello, Alessio and Conti, Francesco and M{\"u}ller, Hanna and Gambardella, Luca Maria and Benini, Luca and Giusti, Alessandro and Guzzi, J{\'e}r{\^o}me},
  journal={IEEE Internet of Things Journal},
  novolume={9},
  nonumber={3},
  nopages={1913--1929},
  year={2021},
  nopublisher={IEEE}
}

@article{zhang2020modular,
      title={A Modular Robotic Arm Control Stack for Research: Franka-Interface and {FrankaPy}}, 
      author={Kevin Zhang and Mohit Sharma and Jacky Liang and Oliver Kroemer},
      year={2020},
      journal={arXiv preprint arXiv:2011.02398},
      nourl={https://arxiv.org/abs/2011.02398}, 
}

@inproceedings{schuitema2010control,
  title={Control delay in reinforcement learning for real-time dynamic systems: A memoryless approach},
  author={Schuitema, Erik and Bu{\c{s}}oniu, Lucian and Babu{\v{s}}ka, Robert and Jonker, Pieter},
  booktitle={IEEE/RSJ International Conference on Intelligent Robots and Systems (IROS)},
  nopages={3226--3231},
  year={2010},
  noorganization={IEEE}
}

@inproceedings{tan2018sim,
  author       = {Jie Tan and
                  Tingnan Zhang and
                  Erwin Coumans and
                  Atil Iscen and
                  Yunfei Bai and
                  Danijar Hafner and
                  Steven Bohez and
                  Vincent Vanhoucke},
  title        = {Sim-to-Real: Learning Agile Locomotion For Quadruped Robots},
  booktitle    = RSS,
  year         = {2018},
  nourl          = {http://www.roboticsproceedings.org/rss14/p10.html},
  nodoi          = {10.15607/RSS.2018.XIV.010},
}

@inproceedings{dimitrova2020towards,
  title={Towards low-latency high-bandwidth control of quadrotors using event cameras},
  author={Dimitrova, Rika Sugimoto and Gehrig, Mathias and Brescianini, Dario and Scaramuzza, Davide},
  booktitle=ICRA,
  nopages={4294--4300},
  year={2020},
  noorganization={IEEE}
}

@inproceedings{kalashnikov2018scalable,
  title={Scalable deep reinforcement learning for vision-based robotic manipulation},
  author={Kalashnikov, Dmitry and Irpan, Alex and Pastor, Peter and Ibarz, Julian and Herzog, Alexander and Jang, Eric and Quillen, Deirdre and Holly, Ethan and Kalakrishnan, Mrinal and Vanhoucke, Vincent and others},
  booktitle={Conference on Robot Learning (CORL)},
  nopages={651--673},
  year={2018},
  noorganization={PMLR}
}

@article{ivanov2025neural,
  title={Neural network compression for reinforcement learning tasks},
  author={Ivanov, Dmitry A and Larionov, Denis A and Maslennikov, Oleg V and Voevodin, Vladimir V},
  journal={Scientific Reports},
  novolume={15},
  nonumber={1},
  nopages={9718},
  year={2025},
  nopublisher={Nature Publishing Group UK London}
}

@article{degrave2022magnetic,
  title={Magnetic control of tokamak plasmas through deep reinforcement learning},
  author={Degrave, Jonas and Felici, Federico and Buchli, Jonas and Neunert, 
Michael and Tracey, Brendan and Carpanese, Francesco and Ewalds, Timo 
and Hafner, Roland and Abdolmaleki, Abbas and de las Casas, Diego and 
Donner, Craig and Fritz, Leslie and Galperti, Cristian and Huber, 
Andrea and Keeling, James and Tsimpoukelli, Maria and Kay, Jackie and 
Merle, Antoine and Moret, Jean-Marc and Noury, Seb and Pesamosca, 
Federico and Pfau, David and Sauter, Olivier and Sommariva, Cristian 
and Coda, Stefano and Duval, Basil and Fasoli, Ambrogio and Kohli, 
Pushmeet and Kavukcuoglu, Koray and Hassabis, Demis and Riedmiller, 
Martin},
  journal={Nature},
  novolume={602},
  nonumber={7897},
  nopages={414--419},
  year={2022},
  nopublisher={Nature Publishing Group UK London}
}

@article{chevtchenko2021combining,
  title={Combining {STDP} and binary networks for reinforcement learning from images and sparse rewards},
  author={Chevtchenko, S{\'e}rgio F and Ludermir, Teresa B},
  journal={Neural Networks},
  novolume={144},
  nopages={496--506},
  year={2021},
  nopublisher={Elsevier}
}

@book{barto2021reinforcement,
    author = {Sutton, Richard S. and Barto, Andrew G.},
    title = {Reinforcement Learning: An Introduction},
    year = {2018},
    publisher = {MIT Press},
}

@article{huang2022cleanrl,
  author  = {Shengyi Huang and Rousslan Fernand Julien Dossa and Chang Ye and Jeff Braga and Dipam Chakraborty and Kinal Mehta and João G.M. Araújo},
  title   = {{CleanRL}: High-quality Single-file Implementations of Deep Reinforcement Learning Algorithms},
  journal = {Journal of Machine Learning Research (JMLR)},
  year    = {2022},
  novolume  = {23},
  nonumber  = {274},
  nopages   = {1--18},
  nourl     = {http://jmlr.org/papers/v23/21-1342.html}
}

@inproceedings{sac,
  title={Soft actor-critic: Off-policy maximum entropy deep reinforcement learning with a stochastic actor},
  author={Haarnoja, Tuomas and Zhou, Aurick and Abbeel, Pieter and Levine, Sergey},
  booktitle=ICML,
  nopages={1861--1870},
  year={2018},
  noorganization={Pmlr}
}

@inproceedings{tan2022rlx2,
  title={{RL}x2: Training a sparse deep reinforcement learning model from scratch},
  author={Tan, Yiqin and Hu, Pihe and Pan, Ling and Huang, Jiatai and Huang, Longbo},
  booktitle=ICRA,
  year={2023}
}

@article{kadokawa2021binarized,
  title={Binarized P-network: Deep reinforcement learning of robot control from raw images on {FPGA}},
  author={Kadokawa, Yuki and Tsurumine, Yoshihisa and Matsubara, Takamitsu},
  journal={IEEE Robotics and Automation Letters},
  novolume={6},
  nonumber={4},
  nopages={8545--8552},
  year={2021},
  nopublisher={IEEE}
}

@inproceedings{valencia2019using,
  title={Using neuroevolved binary neural networks to solve reinforcement learning environments},
  author={Valencia, Raul and Sham, Chiu-Wing and Sinnen, Oliver},
  booktitle={IEEE Asia Pacific Conference on Circuits and Systems (APCCAS)},
  nopages={301--304},
  year={2019},
  noorganization={IEEE}
}

@InProceedings{ma2025network,
  title = 	 {Network Sparsity Unlocks the Scaling Potential of Deep Reinforcement Learning},
  author =       {Ma, Guozheng and Li, Lu and Wang, Zilin and Shen, Li and Bacon, Pierre-Luc and Tao, Dacheng},
  booktitle = 	 ICML,
  nopages = 	 {42009--42029},
  year = 	 {2025},
  noeditor = 	 {Singh, Aarti and Fazel, Maryam and Hsu, Daniel and Lacoste-Julien, Simon and Berkenkamp, Felix and Maharaj, Tegan and Wagstaff, Kiri and Zhu, Jerry},
  novolume = 	 {267},
  noseries = 	 {Proceedings of Machine Learning Research},
  nomonth = 	 {13--19 Jul},
  nopublisher =    {PMLR},
  nopdf = 	 {https://raw.githubusercontent.com/mlresearch/v267/main/assets/ma25l/ma25l.pdf},
  nourl = 	 {https://proceedings.mlr.press/v267/ma25l.html},
}

@article{arnob2024efficient,
  title={Efficient reinforcement learning by discovering neural pathways},
  author={Arnob, Samin Yeasar and Ohib, Riyasat and Plis, Sergey and Zhang, Amy and Sordoni, Alessandro and Precup, Doina},
  journal=NeurIPS,
  novolume={37},
  nopages={18660--18694},
  year={2024}
}

@article{krishnan2019quarl,
  title={Qua{RL}: Quantization for fast and environmentally sustainable reinforcement learning},
  author={Krishnan, Srivatsan and Lam, Maximilian and Chitlangia, Sharad and Wan, Zishen and Barth-Maron, Gabriel and Faust, Aleksandra and Reddi, Vijay Janapa},
  journal={Transactions on Machine Learning Research (TMLR)},
  year={2022}
}

@inproceedings{graesser2022state,
  title={The state of sparse training in deep reinforcement learning},
  author={Graesser, Laura and Evci, Utku and Elsen, Erich and Castro, Pablo Samuel},
  booktitle=ICML,
  nopages={7766--7792},
  year={2022},
  noorganization={PMLR}
}

@article{lu2024impact,
      title={The Impact of Quantization and Pruning on Deep Reinforcement Learning Models}, 
      author={Heng Lu and Mehdi Alemi and Reza Rawassizadeh},
      year={2024},
      journal={arXiv preprint arXiv:2407.04803},
      nourl={https://arxiv.org/abs/2407.04803}, 
}

@InProceedings{bjork2021lowprecisionrl,
  title = 	 {Low-Precision Reinforcement Learning: Running Soft Actor-Critic in Half Precision},
  author =       {Bj{\"o}rck, Johan and Chen, Xiangyu and De Sa, Christopher and Gomes, Carla P and Weinberger, Kilian},
  booktitle = 	ICML,
  nopages = 	 {980--991},
  year = 	 {2021},
  noeditor = 	 {Meila, Marina and Zhang, Tong},
  novolume = 	 {139},
  noseries = 	 {Proceedings of Machine Learning Research},
  nomonth = 	 {18--24 Jul},
  nopublisher =    {PMLR},
  nopdf = 	 {http://proceedings.mlr.press/v139/bjorck21a/bjorck21a.pdf},
  nourl = 	 {https://proceedings.mlr.press/v139/bjorck21a.html}
}

@inproceedings{han2015learning,
  title={Learning both weights and connections for efficient neural network},
  author={Han, Song and Pool, Jeff and Tran, John and Dally, William},
  booktitle=NeurIPS,
  nvolume={28},
  year={2015}
}

@article{rati2025adaptive,
  title={Adaptive Dataflow and Precision Optimization for Deep Learning on Configurable Hardware Architectures},
  author={Rati, Gulnaz and Mendes, Rafael and Noor, Aisha},
  year={2025}
}

@article{bjorck2021towards,
  title={Towards deeper deep reinforcement learning with spectral normalization},
  author={Bjorck, Nils and Gomes, Carla P and Weinberger, Kilian Q},
  journal=NeurIPS,
  nvolume={34},
  npages={8242--8255},
  year={2021}
}

@inproceedings{dadashi2021continuous,
  title={Continuous control with action quantization from demonstrations},
  author={Dadashi, Robert and Hussenot, L{\'e}onard and Vincent, Damien and Girgin, Sertan and Raichuk, Anton and Geist, Matthieu and Pietquin, Olivier},
  booktitle=ICML,
  year={2022}
}

@inproceedings{grossman2022just,
  author = {Grossman, Lev and Plancher, Brian},
  title = {Just Round: Quantized Observation Spaces Enable Memory Efficient Learning of Dynamic Locomotion},
  booktitle=ICRA,
  noaddress = {London, UK},
  nomonth={May.},
  year = {2023}
}

@article{bengio2013estimating,
      title={Estimating or Propagating Gradients Through Stochastic Neurons for Conditional Computation}, 
      author={Yoshua Bengio and Nicholas Léonard and Aaron Courville},
      year={2013},
      journal={arXiv preprint arXiv:1308.342},
      nourl={https://arxiv.org/abs/arXiv:1308.3432}, 
}

@inproceedings{jacob2018quantization,
  title={Quantization and training of neural networks for efficient integer-arithmetic-only inference},
  author={Jacob, Benoit and Kligys, Skirmantas and Chen, Bo and Zhu, Menglong and Tang, Matthew and Howard, Andrew and Adam, Hartwig and Kalenichenko, Dmitry},
  booktitle=CVPR,
  nopages={2704--2713},
  year={2018}
}

@article{gil2021quantization,
  title={Quantization-aware pruning criterion for industrial applications},
  author={Gil, Yoonhee and Park, Jong-Hyeok and Baek, Jongchan and Han, Soohee},
  journal={IEEE Transactions on Industrial Electronics},
  novolume={69},
  nonumber={3},
  nopages={3203--3213},
  year={2021},
  nopublisher={IEEE}
}

@article{blott2018finn,
  title={{FINN-R}: An end-to-end deep-learning framework for fast exploration of quantized neural networks},
  author={Blott, Michaela and Preu{\ss}er, Thomas B and Fraser, Nicholas J and Gambardella, Giulio and O’brien, Kenneth and Umuroglu, Yaman and Leeser, Miriam and Vissers, Kees},
  journal={ACM Transactions on Reconfigurable Technology and Systems (TRETS)},
  novolume={11},
  nonumber={3},
  nopages={1--23},
  year={2018},
  nopublisher={ACM New York, NY, USA}
}

@inproceedings{finn,
author = {Umuroglu, Yaman and Fraser, Nicholas J. and Gambardella, Giulio and Blott, Michaela and Leong, Philip and Jahre, Magnus and Vissers, Kees},
title = {{FINN}: A Framework for Fast, Scalable Binarized Neural Network Inference},
booktitle = {ACM/SIGDA International Symposium on Field-Programmable Gate Arrays},
noseries = {FPGA '17},
year = {2017},
nopages = {65--74},
nopublisher = {ACM}
}

@article{lazarus2022deepbinaryreinforcementlearning,
      title={Deep Binary Reinforcement Learning for Scalable Verification}, 
      author={Christopher Lazarus and Mykel J. Kochenderfer},
      year={2022},
      journal={arXiv preprint arXiv:2203.05704},
}

\newpage

\appendix
\section{Hyperparameters}
\label{app:hyperparameters}

Default models use the ReLU activation function and a hidden layer width of $256$ neurons.
\begin{table}[h!]
\centering
\small
\caption{SAC hyperparameters used in our experiments.}
\label{tab:sac_hparams}
\begin{tabular}{@{}llp{6.8cm}@{}}
\toprule
\textbf{Hyperparameter} & \textbf{Value} & \textbf{Notes} \\
\midrule
Total timesteps & $1{,}000{,}000$ & Environment steps during training. \\
Replay buffer size & $1\times 10^{6}$ & Transitions stored. \\
Discount $\gamma$ & 0.99 & Future reward discount. \\
Target smoothing $\tau$ & 0.005 & Soft target update rate. \\
Batch size & 256 & Minibatch size for updates. \\
Learning starts & $5\times 10^{3}$ & Steps collected before learning. \\
Policy LR & $3\times 10^{-4}$ & Adam LR for policy. \\
Q-network LR & $1\times 10^{-3}$ & Adam LR for critics. \\
Policy update frequency & 2 & Update policy every 2 critic steps. \\
Target network frequency & 1 & Target update every critic step. \\
Entropy & autotune & Adapted online. \\
\bottomrule
\end{tabular}
\end{table}
\begin{table}[h!]
\centering
\small
\caption{DDPG hyperparameters used in our experiments.}
\label{tab:ddpg_hparams}
\begin{tabular}{@{}llp{6.8cm}@{}}
\toprule
\textbf{Hyperparameter} & \textbf{Value} & \textbf{Notes} \\
\midrule
Total timesteps & $1{,}000{,}000$ & Environment steps during training. \\
Learning rate & $3\times 10^{-4}$ & Optimizer step size (actor \& critic). \\
Replay buffer size & $1\times 10^{6}$ & Transitions stored. \\
Discount $\gamma$ & 0.99 & Future reward discount. \\
Target smoothing $\tau$ & 0.005 & Smoothing for target nets. \\
Batch size & 256 & Minibatch size for critic/actor updates. \\
Exploration noise (std) & 0.1 & Gaussian action noise during data collection. \\
Learning starts & $2.5\times 10^{4}$ & Steps collected before learning. \\
Policy update frequency & 2 & Update policy every 2 critic steps. \\
\bottomrule
\end{tabular}
\end{table}

\section{Model Selection}

\label{app:ms}
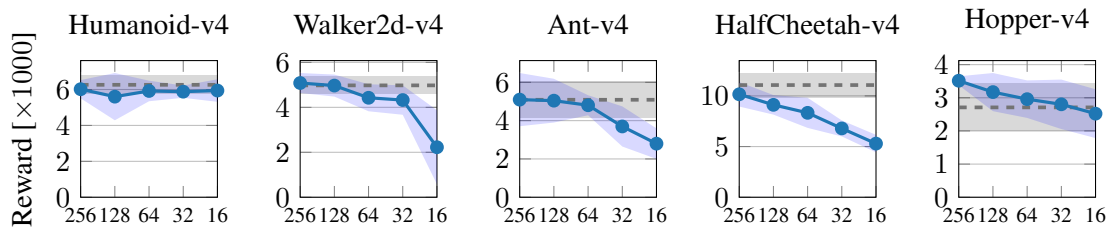
\begin{figure}[H]
    \centering
 \begin{tikzpicture}    
    \begin{groupplot}[
      group style={
        group size=5 by 1,          %
        horizontal sep=1.1cm,
        vertical sep=1.3cm,
      },
      sizeaxis,                     %
      legend columns=4,
      legend to name=globalLegend3,
      xmin=16, xmax=256,
      log basis x=2,
      xmode=log,
      xtick={16, 32, 64, 128, 256},
      xticklabels={16, 32, 64, 128, 256},
      ymin=0,
      scaled y ticks=false,
      ylabel={Reward [$\times 1000$]},
      yticklabel={\pgfmathparse{\tick/1000}\pgfmathprintnumber{\pgfmathresult}}, 
      xticklabel style={font=\scriptsize}
    ]
    \nextgroupplot[title={Humanoid-v4}]
      \sizepanel{data/sac/sizes}{Humanoid-v4}{3}
              \nextgroupplot[title={Walker2d-v4}, ylabel={}]
      \sizepanel{data/sac/sizes}{Walker2d-v4}{2}
          \nextgroupplot[title={Ant-v4}, ylabel={}]
      \sizepanel{data/sac/sizes}{Ant-v4}{2}

              \nextgroupplot[title={HalfCheetah-v4},ylabel={}]
      \sizepanel{data/sac/sizes}{HalfCheetah-v4}{3}

    \nextgroupplot[title={Hopper-v4}, ylabel={}]
      \sizepanel{data/sac/sizes}{Hopper-v4}{2}

    \end{groupplot}
 \end{tikzpicture}
 
 \caption{Return vs. hidden width for SAC under the minimal FP32-matching \textbf{core} precision (2-bit except 3-bit for HalfCheetah/Humanoid). FP32 mean and its one-standard deviation band shown for reference.}
 \label{fig:size}
\end{figure}

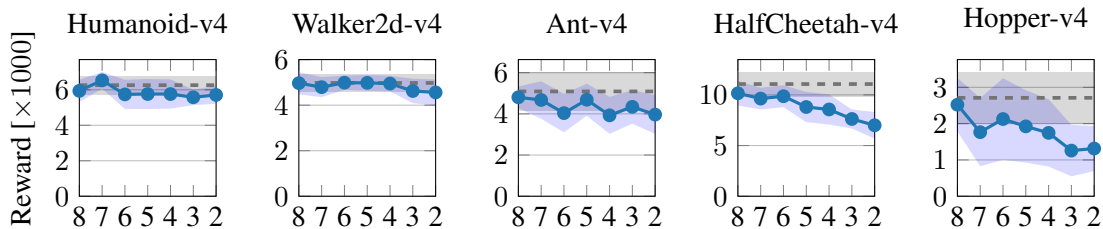
\begin{figure}[H]
    \centering
 \begin{tikzpicture}    
    \begin{groupplot}[
      group style={
        group size=5 by 1,          %
        horizontal sep=1.1cm,
        vertical sep=1.3cm,
      },
      sizeaxis,                     %
      legend columns=4,
      legend to name=globalLegend4,
      xmin=2, xmax=8,
      xtick={8,7,6,5,4,3,2},
      xticklabels={8,7,6,5,4,3,2},
      ylabel={Reward [$\times 1000$]},
      ymin=0,
      scaled y ticks=false,
      yticklabel={\pgfmathparse{\tick/1000}\pgfmathprintnumber{\pgfmathresult}}, 
      ymin=0,
    ]
    \nextgroupplot[title={Humanoid-v4}]
      \inputpanel{data/sac/input_activations}{Humanoid-v4}{16}{3}
              \nextgroupplot[title={Walker2d-v4}, ylabel={}]
      \inputpanel{data/sac/input_activations}{Walker2d-v4}{128}{2}
          
    \nextgroupplot[title={Ant-v4}, ylabel={}]
      \inputpanel{data/sac/input_activations}{Ant-v4}{64}{2}
              \nextgroupplot[title={HalfCheetah-v4},ylabel={}]
\inputpanel{data/sac/input_activations}{HalfCheetah-v4}{256}{3}

    \nextgroupplot[title={Hopper-v4}, ylabel={}]
      \inputpanel{data/sac/input_activations}{Hopper-v4}{16}{2}

    \end{groupplot}
 \end{tikzpicture}
  \caption{Return vs. input quantization for SAC under the configuration from Table \ref{tab:final-config}, except input bits, which is swept here. FP32 mean and its one-standard deviation band shown for reference.}
 \label{fig:input}
\end{figure}

\section{Input Normalization}

\label{app:normalization}
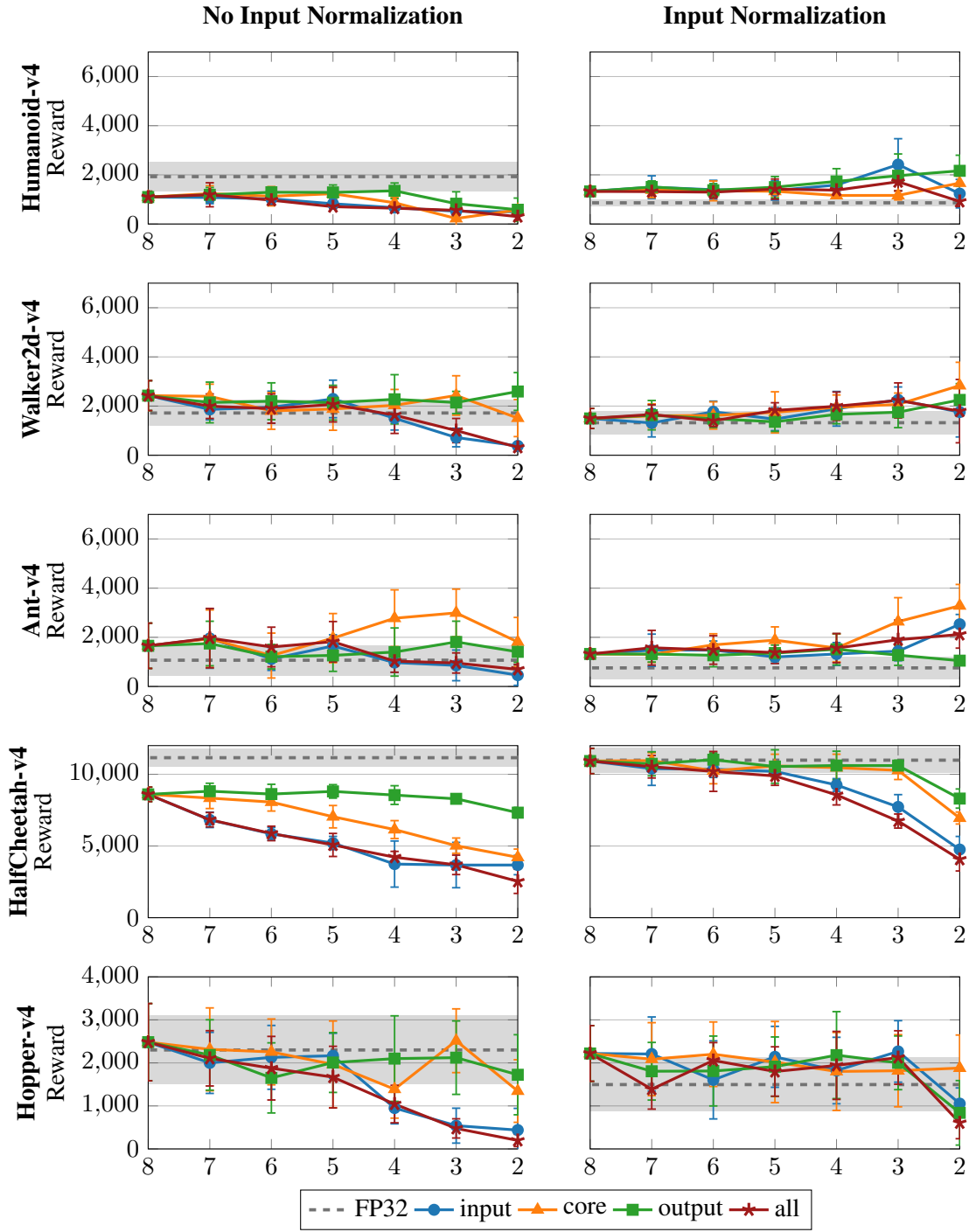
\begin{figure}
\begin{tikzpicture}
\begin{groupplot}[
  group style={group size=2 by 5, horizontal sep=1.1cm, vertical sep=0.8cm},
  myaxis,
  legend to name=globalLegend, legend columns=5,
]

\nextgroupplot[title={\textbf{No Input Normalization}},ymin=0, ymax=7000,ylabel=\shortstack{\textbf{Humanoid-v4}\\Reward}]
\envpanel{data/ddpg_old_0811}{Humanoid-v4}

\nextgroupplot[title={\textbf{Input Normalization}},ymin=0, ymax=7000,yticklabels=\empty, ylabel={},]
\envpanel{data/ddpg_new}{Humanoid-v4}

\nextgroupplot[title={},ymin=0, ymax=7000,ylabel=\shortstack{\textbf{Walker2d-v4}\\Reward}]
\envpanel{data/ddpg_old_0811}{Walker2d-v4}

\nextgroupplot[title={},ymin=0, ymax=7000,yticklabels=\empty, ylabel={},]
\envpanel{data/ddpg_new}{Walker2d-v4}

\nextgroupplot[title={}, ymin=0, ymax=7000,ylabel=\shortstack{\textbf{Ant-v4}\\Reward}]
\envpanel{data/ddpg_old_0811}{Ant-v4}

\nextgroupplot[title={},ymin=0, ymax=7000,yticklabels=\empty, ylabel={}]
\envpanel{data/ddpg_new}{Ant-v4}

\nextgroupplot[title={},ymin=0, ymax=12000,ylabel=\shortstack{\textbf{HalfCheetah-v4}\\Reward}]
\envpanel{data/ddpg_old_0811}{HalfCheetah-v4}

\nextgroupplot[title={},ymin=0, ymax=12000,yticklabels=\empty, ylabel={},]
\envpanel{data/ddpg_new}{HalfCheetah-v4}

\nextgroupplot[title={ },ymin=0, ymax=4000,ylabel=\shortstack{\textbf{Hopper-v4}\\Reward}]
\envpanel{data/ddpg_old_0811}{Hopper-v4}

\nextgroupplot[title={},ymin=0, ymax=4000,yticklabels=\empty,ylabel={}]
\envpanel{data/ddpg_new}{Hopper-v4}

\end{groupplot}
\node at (6.2,-14.5) {\pgfplotslegendfromname{globalLegend}};
\end{tikzpicture}
\caption{DDPG; {Reward \vs bitwidth} for full-precision (FP32) baselines (shaded region indicates one standard deviation) as well as four variants of network quantization: \emph{all}: all network operations are quantized to indicated bitwidths; \emph{input}/\emph{output}: the quantization of only the inputs/outputs are varied; \emph{core}: the quantization of weights and internal activations are varied. In the latter three cases, all other components are left at 8-bit precision. We achieve FP32-parity on most task bitwidth combinations with DDPG when we utilize input normalization.}
\label{fig:normalized_vsnot}
\end{figure}

\textbf{SAC} For SAC, we find that running input normalization improves returns for the floating-point baseline (except HalfCheetah, but results are close with standard deviations overlapping); see Table \ref{tab:normalizedvsnotSAC}. Due to this, and SAC's better overall returns, when compared to DDPG, we generally apply normalization across this work. {We hypothesize that input normalization improves returns, specifically in the quantization context, because it improves optimization of the quantization scale parameter in the first layer, which we show to have major impact on returns in Section \ref{sec:experiments}.} \\
\begin{table}[h!]
\centering
\begin{tabular}{l l l}
\toprule
Environment &
\multicolumn{1}{c}{No Input Normalization} &
\multicolumn{1}{c}{Input Normalization} \\
\midrule
Ant         & \numk{4974.88}    $\pm$ \numk{1738.89}      & \numk{5089.34372}  $\pm$ \numk{943.0451170820312} \\
Hopper      & \numk{2400.29}    $\pm$ \numk{952.75}       & \numk{2710.48154}  $\pm$ \numk{711.6178800368248} \\
HalfCheetah & \numk{11562.85}   $\pm$ \numk{1011.04}      & \numk{11066.0692}  $\pm$ \numk{1190.0712830394095} \\
Walker2d    & \numk{4316.83}    $\pm$ \numk{929.06}       & \numk{4976.8076}   $\pm$ \numk{395.3054581626157} \\
Humanoid    & \numk{4671.71}    $\pm$ \numk{1421.12}      & \numk{6261.65173}  $\pm$ \numk{512.5055959710014} \\
\bottomrule
\end{tabular}
\caption{Floating point baseline, with and without input normalization. Mean and standard deviation over 10 trained models and 1000 rollouts per model. The floating point baseline with input normalization performs on par or better than without normalization.}
\label{tab:normalizedvsnotSAC}
\end{table}
\\
\textbf{DDPG} Figure \ref{fig:normalized_vsnot} shows a floating point baseline and various quantization scopes trained with DDPG, as described in Section \ref{sec:experiments}. The results for the left-hand side plots were obtained \emph{without} input normalization (as implemented in CleanRL), whereas we use running, per-dimension input normalization for the right-hand side plots. While the floating-point baseline with running mean performs worse than its non-normalized counterpart, we observe the opposite behaviour for the quantized networks. Importantly, the quantized networks \emph{with} input normalization consistently perform on par with the floating-point networks \emph{without} input normalization (which are the stronger baselines).

\end{document}